\documentclass[conference,compsoc]{IEEEtran}

\usepackage[usenames,dvipsnames,table,xcdraw]{xcolor}

\usepackage{tcolorbox}
\tcbuselibrary{skins}
\newtcolorbox{answerbox}{
  colback=black!5!white,
  colframe=black!70!white,
  fonttitle=\bfseries,
  sharp corners,
  rounded corners,
  boxrule=0.3mm
}

\usepackage{pifont}
\newcommand{\cmark}{\ding{51}}
\newcommand{\xmark}{\ding{55}}

\usepackage{listings}
\colorlet{keyword}{blue!100!black!80}
\colorlet{STD}{Lavender}
\colorlet{comment}{green!90!black!90}
\definecolor{mygreen}{rgb}{0,0.6,0}
\definecolor{mygray}{rgb}{0.5,0.5,0.5}
\definecolor{mymauve}{rgb}{0.58,0,0.82}

\lstdefinestyle{C}{ 
  language     = C,
  basicstyle   = \footnotesize \ttfamily,
  keywordstyle = [1]\color{keyword}\bfseries,
  keywordstyle = [2]\color{STD}\bfseries,
  commentstyle = \color{mygreen}\itshape,
  backgroundcolor=\color{white},   
  columns=fullflexible,            
  basicstyle=\footnotesize,        
  breakatwhitespace=false,         
  breaklines=false,                
  captionpos=c,                    
  extendedchars=true,              
  frame=single,                    
  keepspaces=true,                 
  numbers=left,                    
  numbersep=5pt,                   
  numberstyle=\tiny\color{mygray}, 
  rulecolor=\color{black},         
  showspaces=false,                
  showstringspaces=false,          
  showtabs=false,                  
  stepnumber=1,                    
  stringstyle=\color{mymauve},     
  tabsize=2,                       
  title=\lstname,                  
  literate=                        
     {Ö}{{\"O}}1 
     {Ä}{{\"A}}1 
     {Ü}{{\"U}}1 
     {ß}{{\ss}}2 
     {ü}{{\"u}}1 
     {ä}{{\"a}}1 
     {ö}{{\"o}}1 
     {â}{{\^{a}}}1 
     {Â}{{\^{A}}}1 
     {ç}{{\c{c}}}1 
     {Ç}{{\c{C}}}1 
     {ğ}{{\u{g}}}1 
     {Ğ}{{\u{G}}}1 
     {ı}{{\i}}1 
     {İ}{{\.{I}}}1 
     {ş}{{\c{s}}}1 
     {Ş}{{\c{S}}}1 
}
\usepackage{subfig} 
\usepackage{enumitem}
\usepackage{tikz}
\usepackage{svg}
\usetikzlibrary{shapes, positioning, intersections, arrows, automata, positioning, decorations.pathreplacing, decorations.pathmorphing, shapes.gates.logic.US,shapes.gates.logic.IEC, shapes, intersections}
\usepackage{float}

\usepackage{cite}

\usepackage{amsmath}

\usepackage{url}
\usepackage{hyperref}

\usepackage{cleveref}

\usepackage{threeparttable}
\usepackage{booktabs}
\usepackage{tabularx}

\usepackage{titlesec}
\newcommand{\appendixstyle}{
    \titleformat{\section}
    [block] 
    {\centering\LARGE\bfseries} 
    {\appendixname~\thesection} 
    {1em} 
    {} 

    \titleformat{\subsection}
    [block]
    {\centering\Large\bfseries} 
    {\thesubsection} 
    {1em}
    {}
}

\usepackage{orcidlink}

\usepackage{fancyhdr}
\usepackage{graphicx} 
\fancypagestyle{firstpage}{
    \fancyfoot[L]{
        \raisebox{1.9cm}[0pt][0pt]{ 
            \begin{minipage}[t]{\columnwidth}
                \rule{\linewidth}{0.4pt} 
                \textbf{Accepted at Applied Cryptography and Network \\ Security (ACNS) 2025}
            \end{minipage}
        }
    }
    \fancyfoot[C]{} 
    \fancyfoot[R]{} 
}

\usepackage[font=it]{caption}
\begin{document}

\title{
OCEAN: Open-World Contrastive Authorship Identification
}
\author{
    \IEEEauthorblockN{Felix Mächtle\,\orcidlink{0009-0009-2431-0322},
    Jan-Niclas Serr\,\orcidlink{0009-0001-4624-8592},
    Nils Loose\,\orcidlink{0009-0003-6243-1623},
    Jonas Sander\,\orcidlink{0009-0007-9402-823X},
    Thomas Eisenbarth\,\orcidlink{0000-0003-1116-6973}}
    \IEEEauthorblockA{Institute for IT Security, Luebeck, Germany \\        
    Email: \{f.maechtle, j.serr, n.loose, j.sander, thomas.eisenbarth\}@uni-luebeck.de}
}

\renewcommand{\tablename}{Table}

\def\UrlBreaks{\do\/\do-}

\newcommand{\etal}[1]{#1 \emph{et al.}}
\newcommand{\schemename}{\textsc{OCEAN}}
\newcommand{\dataset}{\textsc{Conan}}
\newcommand{\snoopy}{\textsc{Snoopy}}

\maketitle

\begin{abstract}
In an era where cyberattacks increasingly target the software supply chain, the ability to accurately attribute code authorship in binary files is critical to improving cybersecurity measures. 
We propose \schemename{}, a contrastive learning-based system for function-level authorship attribution. \schemename{} is the first framework to explore code authorship attribution on compiled binaries in an open-world and extreme scenario, where two code samples from unknown authors are compared to determine if they are developed by the same author. 
To evaluate OCEAN, we introduce new realistic datasets: \dataset{}, to improve the performance of authorship attribution systems in real-world use cases, and \snoopy{}, to increase the robustness of the evaluation of such systems. 
We use \dataset{} to train our model and evaluate on \snoopy{}, a fully unseen dataset, resulting in an AUROC score of 0.86 even when using high compiler optimizations.
We further show that \dataset{} improves performance by 7\% compared to the previously used Google Code Jam dataset.
Additionally, \schemename{} outperforms previous methods in their settings, achieving a 10\% improvement over state-of-the-art SCS-Gan in scenarios analyzing source code. Furthermore, \schemename{} can detect code injections from an unknown author in a software update, underscoring its value for securing software supply chains.
\end{abstract}
\begin{IEEEkeywords}
Authorship Identification, Binary Analysis, Contrastive Learning
\end{IEEEkeywords}

\thispagestyle{firstpage}
\section{Introduction}

According to the World Economic Forum's Global Risk Report 2024, cyberattacks are among the five biggest risks in 2024~\cite{global_risk_report}. In particular, the software supply chain, where externally developed code is introduced into critical parts of a network, represents a rising attack surface~\cite{software_supply_chain_report}. The ability to detect and attribute malicious code is an essential part of protection~\cite{DBLP:Soverein}. Authorship attribution has been extensively studied in fields such as literary criticism, forensic linguistics, and legal proceedings \cite{sari-etal-2018-topic, ai-etal-2022-whodunit, AA_text_survey}. Similar techniques have also been used to identify code authorship as programmers leave stylistic signatures through distinct coding methodologies, preferences, and design patterns. Previous studies have demonstrated the feasibility of identifying programmers through their coding styles in both source code and binary forms~\cite{bin_author, bin_eye, multi_x, scsgan, 10.5555/2831143.2831160, Dauber2017GitBW, hao2022multiauth, DBLP:conf/icse/LiCCZX22, Meng2017BinaryCM, meng_mult_auth, omi2021multiauth, DBLP:journals/access/ZafarSSM20, DBLP:conf/sigsoft/BogomolovKRBB21}.

$\hspace{1cm}$\\

Traditionally, research has viewed the attribution of code authorship as a closed world classification problem. Using techniques ranging from random forests to modern machine learning methods, an unseen or new code sample is attributed to a known author from the training data set. Hence, only a finite set of authors is considered, and sample training data must be available for every author~\cite{meng_mult_auth, omi2021multiauth, multi_x, hao2022multiauth, DBLP:conf/icse/LiCCZX22, Meng2017BinaryCM, DBLP:conf/sigsoft/BogomolovKRBB21}. This inadvertently limits the applicability of the results as an application in a real-world scenario rarely provides a priori knowledge and sample data for every attributable author. Some techniques consider a limited open world scenario that reformulates the attribution as a membership problem, considering whether an author is part of the training dataset or not\cite{Dauber2017GitBW, 10.5555/2831143.2831160, bin_author, bin_eye, DBLP:journals/access/ZafarSSM20}. While this limited open world scenario enables applicability to authors beyond the known authors, it allows only coarse-grained attribution. 

\etal{Ou}~\cite{scsgan} adopt a genuinely open world approach to authorship attribution, where no predefined set of authors is assumed. Instead, they compare two source code extracts to predict whether they originate from the same author. While this approach, termed \textit{extreme authorship identification}, enhances applicability in real-world scenarios, it remains constrained by two limitations. First, their evaluation is restricted to file-level granularity and relies on the original source code. Second, the dataset used for evaluation has been shown to lack representation of realistic production-level code~\cite{dataset, DBLP:conf/sigsoft/BogomolovKRBB21}.

We address this critical gap by proposing a novel approach, \schemename{}, designed to predict whether two functions from a binary file belong to the same author. To ensure a realistic evaluation, we train and assess our method using two newly curated datasets derived from large-scale open-source software projects. Our approach leverages contrastive learning to train \textit{UniXcoder}~\cite{unixcoder}, a state-of-the-art machine learning model. To the best of our knowledge, \schemename{} is the first authorship identification method to operate on binary files within an extreme open world scenario.

Training and evaluation on diverse and realistic data sets is paramount to the trustworthiness of reported performance and the transferability of results~\cite{DBLP:conf/uss/Arp/dos-and-donts}.  
Hence, we introduce two novel datasets containing \texttt{C} and \texttt{C++} programs. 
We focus on \texttt{C} and \texttt{C++}, as they hold the largest market share (37.2\%) among compiled programming languages,  according to the Stack Overflow Developer Survey 2024 \cite{stackoverflow_survey_2024}.
The first dataset, \dataset{}, consists of 464 annotated binaries created from Conan \cite{conan}, a popular package manager for \texttt{C}/\texttt{C++} libraries, and their corresponding GitHub repositories. Using the history of edits within git, we infer authorship information for each line of code, which we then correlate with its compiled binary output. The \dataset{} dataset is used exclusively for training.
The second dataset, \snoopy{}, a manually curated dataset consisting of seven popular GitHub repositories (Bitcoin~\cite{snoopy-bitcoin}, Curl~\cite{snoopy-curl}, GLibC~\cite{snoopy-glibc}, Nginx~\cite{snoopy-nginx}, OpenSSH~\cite{snoopy-openssh}, PHP~\cite{snoopy-php} and Redis~\cite{snoopy-redis}), is used exclusively for evaluation. 
We use two distinct datasets for training and evaluation as a precaution to prevent data snooping~\cite{DBLP:conf/uss/Arp/dos-and-donts}, a bias we find in several of the preceding studies~\cite{10.5555/2831143.2831160, multi_x, scsgan, meng_mult_auth, omi2021multiauth, Dauber2017GitBW, hao2022multiauth, DBLP:conf/sigsoft/BogomolovKRBB21}. Data snooping occurs when a model learns dataset-specific structural features, leading to biased evaluations and inflated performances. Using a fully unseen dataset we assure robust performance results.

In our comprehensive evaluation, we outperform \textit{SCS-Gan}, the previous state-of-the-art by \etal{Ou}~\cite{scsgan}, by 10\% in their original scenario where the original source code is available. In our more challenging scenario, applied to a fully unseen dataset of binaries, our approach achieves an AUROC score of 0.86. Additionally, in a case study we demonstrate the ability of \schemename{} to successfully identify new authors between two versions of a binary. 
We also found that training with our \dataset{} dataset yields a 7\% improvement in realistic scenarios compared to traditional datasets. 
To support future research, we release \schemename{}, \dataset{}, \snoopy{} and our dataset collection tool as open source on GitHub (\url{https://github.com/UzL-ITS/OCEAN}).

\noindent To summarize, our contributions are:
\begin{enumerate}[noitemsep,topsep=0pt,leftmargin=*]

    \item \schemename{} outperforms current state-of-the-art SCS-Gan~\cite{scsgan} by 10\% in their much simpler scenario where the original source code is available.

    \item To the best of our knowledge, \schemename{} is the first framework to explore code authorship attribution on compiled binaries in an open and extreme scenario, achieving an AUROC score of 0.86.

    \item We create a new dataset, \dataset{}, for code attribution and show that it improves performance by $7\%$ compared to the previously used Google Code Jam dataset. In addition, we generate a second dataset, \snoopy{}, to enable unbiased evaluations. 
    
    \item We further demonstrate the successful use of \schemename{} to identify a new author between two versions of a binary, highlighting the potential use of \schemename{} to protect the software supply chain.
    
\end{enumerate}

\section{Background}

\subsection{Extreme Authorship Identification}

The goal of extreme authorship identification is to determine whether two given functions, $x_1$ and $x_2$, were written by the same developer~\cite{scsgan}.
Formally, given two program functions, $x_1$ and $x_2$, where $x_1$ is written by author $a_1$ and $x_2$ by $a_2$,
the objective is to determine whether $a_1 = a_2$. Consequently, the goal is to develop a classifier, $c(x_1, x_2) \in [0,1]$, which returns a value indicating the likelihood that $x_1$ and $x_2$ were written by the same author. The classifier should satisfy the condition $c(x_1, x_2) =
\begin{cases}
  1 & \text{if } a_1 = a_2 \\
  0 & \text{otherwise}
\end{cases}$.

\subsection{Program Representations for Authorship Identification}
\label{sec:background:program-representations}

Programs can be represented in many ways, and the choice of representation can significantly impact the effectiveness of the authorship attribution (\Cref{tab:model_vs_repr_full}).

\noindent\textbf{Source code} is widely used in related work due to its high-level nature and the availability of contextual information \cite{multi_x, Dauber2017GitBW, hao2022multiauth, 10.5555/2831143.2831160, DBLP:conf/icse/LiCCZX22, omi2021multiauth, scsgan, DBLP:journals/access/ZafarSSM20, DBLP:conf/sigsoft/BogomolovKRBB21}. The presence of comments and variable names, such as specific naming conventions or commenting styles, can provide additional clues about an author's identity.

\noindent\textbf{Raw binaries} are the most low-level representation and provide little information about the original source code, as this sparse representation introduces significant noise. The amount of noise makes it difficult for classification models to detect author-specific elements, derived from the source code~\cite{DBLP:conf/fps/AlrabaeeSDW16}.

\noindent\textbf{Decompiled raw binaries} of C/C++ code can bridge the gap between high-level and low-level representations. Although decompiled code might appear similar to the original source code, much information, including variable names and comments, is lost during compilation. The decompiler's style also influences the representation, as it follows specific patterns and conventions that may not exactly match the original author's style \cite{bin_author}.

\noindent\textbf{Assembly instructions and P-code} provide intermediate representations that balance between high-level source code and low-level binary data. 
Assembly instructions turn raw binary code into a human-readable syntax that can easily be assembled back into an executable binary.
P-code, or pseudocode, is a form of intermediate code used by certain compilers to facilitate debugging and portability. In the decompiler Ghidra~\cite{ghidra}, there are two flavors of P-code: Raw P-code, which is a direct translation of machine instructions into a generic, low-level set of operations, and high-level P-code, which undergoes further analysis and transformation to aid in decompilation and readability. All representations still include register names and jump addresses, which could affect the performance of the authorship contribution, as they are not necessarily connected to the author's identity and therefore introduce noise.
We mitigate the noise by replacing such occurrences with the fixed string \texttt{HEXSTR}. For example, we adapt a sequence like \texttt{JUMP 0x1a2b} to \texttt{JUMP HEXSTR} and call those adjusted representations cleaned (e.g., assembly (c)).

An example of a single function in all these representations is given in \Cref{appendix:code-representation-examples}.

\subsection{Contrastive Learning}

Contrastive learning is an ML paradigm in which a model learns to differentiate between similar and dissimilar characteristics of data. The goal is to find embeddings of the data that optimally reveal similarities and differences~\cite{DBLP:conf/cvpr/HadsellCL06}. Supervised contrastive learning is a special case, where label information are available. In that case, pairs with the same label (positive pairs) and pairs with different labels (negative pairs) are used together with a contrastive loss function to train a model to minimize the distance metric between embeddings of positive pairs and to maximize distance between negative pairs \cite{DBLP:conf/emnlp/GaoYC21}.

\section{Datasets for Code Attribution}

Many recent code attribution systems \cite{bin_author, bin_eye,10.5555/2831143.2831160, multi_x, scsgan, DBLP:conf/icse/LiCCZX22, DBLP:journals/access/ZafarSSM20, DBLP:conf/sigsoft/BogomolovKRBB21} were evaluated using the Google Code Jam dataset (GCJ)\cite{gcj}, consisting of solutions for small programming tasks from an annual programming challenge held by Google. The GCJ dataset seems appealing for authorship identification as it links multiple source code files to individual authors.
However, critics have pointed out that GCJ does not resemble real-world production code but rather ad-hoc solutions to specific challenges \cite{dataset, DBLP:conf/sigsoft/BogomolovKRBB21}. Moreover, many samples are contaminated and contain auto-generated comments, such as the \texttt{@author} comment, which make authorship attribution trivial and prevent a robust evaluation of such systems for harsher real-world settings without such information. Other C/C++ datasets for code attribution often share similar problems or are, like the recent dataset of \etal{Li} \cite{DBLP:conf/icse/LiCCZX22}, rather small, limiting the robustness of the evaluation and impede the effective use of data-hungry state-of-the-art ML techniques. 
Given these limitations, a dataset comprising real-world production code is essential for a more accurate evaluation of authorship identification systems.

\begin{table}[t]
    \setlength\tabcolsep{3.4pt}
    \centering
    \begin{threeparttable}
    \caption{Statistics of the used datasets if generated in default settings as described in \Cref{sec:eval:experiment_setup} as well as GCJ for all years.}
    \begin{tabularx}{\columnwidth}{Xrrr}
    \toprule
     & \dataset & \snoopy & GCJ \\
     \midrule
         Unique functions & 42 134 & 9 458 & $\hspace{2pt}$ 512 989 \\
         Authors & 1254             & 344       & 46 223\\ 
         Projects & 464            & 7 & - \\ 
         Lines per function 
            & 61.04 
            & 52.82 
            & 27.63 \\ 
         Authors per function 
            & 1.54 
            & 2.19 
            & 1 \\ 
         Functions per author 
            & 33.60      
            & 27.49 
            & 11.10 \\ 
    \bottomrule 
    \end{tabularx}
    \label{tab:main:conan_dataset_stats}
    \end{threeparttable}
\end{table}

\subsection{New Real World Datasets}\label{ssec:Conan}\label{sec:datasets:generation-of-dataset}
We present two new datasets (see \Cref{tab:main:conan_dataset_stats}) to address the challenges in the robust evaluation of code authorship attribution systems and leverage them for the development of \schemename. We hope these datasets will facilitate the robust evaluation of future code attribution systems and enable fair comparisons of different approaches.

\noindent\textbf{\dataset:} We leverage the \texttt{Conan Package Manager}~\cite{conan} to construct our new dataset called \dataset. Similar to \texttt{Maven} or \texttt{pip}, the \texttt{Conan Package Manager} allows the inclusion of various libraries. Its default build routine can be modified to include custom flags while ensuring successful compilation of all libraries. For each library, we fetched the linked GitHub project and removed header-only and dynamically linked libraries to obtain complete and standalone binaries. 
\dataset{} consists of 464 projects of which 255 are written in C and 209 in C++.

\noindent\textbf{\snoopy:} 
Prior work showed that relying on a single dataset can introduce data snooping~\cite{DBLP:conf/uss/Arp/dos-and-donts}, a common flaw in authorship identification evaluations~\cite{10.5555/2831143.2831160, multi_x, scsgan, meng_mult_auth, omi2021multiauth, Dauber2017GitBW, hao2022multiauth, DBLP:conf/sigsoft/BogomolovKRBB21}. Data snooping occurs when a model learns dataset-specific structural hints, leading to biased evaluations. To mitigate this risk, we manually collected and compiled additional projects from GitHub, ensuring a diverse and unbiased dataset. Specifically, we chose Bitcoin~\cite{snoopy-bitcoin}, Curl~\cite{snoopy-curl}, GLibC~\cite{snoopy-glibc}, Nginx~\cite{snoopy-nginx}, OpenSSH~\cite{snoopy-openssh}, PHP~\cite{snoopy-php} and Redis~\cite{snoopy-redis} as control dataset. 
We call this dataset \snoopy{} as it prevents data snooping.
Utilizing \snoopy{} ensures the evaluation of authorship attribution systems to be robust and generalizable.

\noindent \textbf{Labeling and pair construction:}
To annotate ownership information, we compiled all binaries with debug flags, which ensures that all function names remain in the binary. Those names allow us to uniquely identify and map each binary function to its corresponding source code, provided the function names are unique. In cases where function names are not unique, we discard them.

For each compiled binary, we use Ghidra's scripting API to extract all functions. Ghidra is an open-source reverse engineering tool developed by the NSA that provides extensive capabilities for analyzing binary code~\cite{ghidra}. Using its scripting API, we created custom scripts to automatically extract functions across all representations described in \Cref{sec:background:program-representations}. 
The extracted function names are then matched to the original source code in the respective Git project, which is subsequently added to the dataset as an additional representation.

To attribute ownership to each function in the project, we employed the \texttt{git-author} tool by \etal{Meng}~\cite{git_author}. This tool analyzes the git history of the source code to attribute ownership. By examining the commit history, it determines which lines of code were contributed by which authors. The tool aggregates these information and records the ownership as a percentage that reflects each author's contribution to each function.

\section{\schemename{}}

\label{sec:system-description}

\begin{table}[t]
\centering
\begin{threeparttable}
    \caption{OCEAN in comparison to previous approaches.}
    \begin{tabularx}{\columnwidth}{X@{}cl@{\hskip 8pt}l@{\hskip 8pt}c@{ \hskip 3pt}l}
    \toprule
          Scheme & 
          Binary &
         Language & 
         Scenario & 
         Extreme & Granularity \\
    \midrule
        RoPGen \cite{DBLP:conf/icse/LiCCZX22} & \xmark & C, J & CW & \xmark & File \\
\cite{DBLP:conf/sigsoft/BogomolovKRBB21} & \xmark & C, J, P & CW & \xmark & Function \\
Multi-{\(\chi\)} \cite{multi_x} & \xmark & C & CW & \xmark & Line \\
\cite{hao2022multiauth} & \xmark & C, J & CW & \xmark & Line \\
\cite{omi2021multiauth} & \xmark & J & CW & \xmark & Line \\
BinEye \cite{bin_eye} & \cmark & C & CW & \xmark & File \\
\cite{Meng2017BinaryCM} & \cmark & C & CW & \xmark & Basic Block \\
\cite{meng_mult_auth} & \cmark & C & CW & \xmark & Basic Block \\
\cite{10.5555/2831143.2831160} & \xmark & C, P & OW- & \xmark & File \\
\cite{DBLP:journals/access/ZafarSSM20} & \xmark & C, J, P & OW- & \xmark & File \\
\cite{Dauber2017GitBW} & \xmark & C & OW- & \xmark & Function \\
BinAuthor \cite{bin_author} & \cmark & C & OW- & \xmark & File\\
SCS-Gan \cite{scsgan} & \xmark & C, J, P & OW & \cmark & File \\
\schemename{} & \cmark & C & OW & \cmark & Function\\\bottomrule
    \end{tabularx}
    \begin{tablenotes} \scriptsize
    \item[a] Binary: The solution supports the authorship attribution on compiled binaries without access to the source code.
    \item[b] Language: Shows the supported programming languages: C/C++ (C), Java (J), and Python (P).
    \item[c] Scenario: Distinguishes whether all potential authors are known beforehand (Closed World, CW) or if the system can identify unknown authors (Open World, OW). OW- indicates a semi-Open World where the system recognizes an unknown author without distinguishing between multiple unknowns.
    \item[d] If the approach supports the extreme case, only one reference file of an author is needed for the attribution.
    \item[e] Granularity: Indicates the code unit attributed to an author. If the granularity is below function-level, several code units belonging to the same author are usually merged for attribution.
  \end{tablenotes}
    \label{tab:rel_work_methods}
    \end{threeparttable}
\end{table} 

In this section, we present \schemename{}, a fully automated framework for function-level authorship identification in C/C++ binaries. \schemename{} leverages the \dataset{} dataset, contrastive learning, and supports various representation extraction techniques to build a robust classifier capable of distinguishing authors based on their coding style.
As shown in \Cref{tab:rel_work_methods}, previous work mainly focused on the closed world scenario, or a simplified variant of the much harder open world scenario, a membership inference against the training set. \etal{Ou}~\cite{scsgan} where the first to perform a fully open world scenario, however their system does not allow binary analysis. 
\schemename{} is the first system that supports binary analysis in the extreme open world scenario. Additionally \schemename{} allows for a higher granularity than \etal{Ou} as it attributes functions instead of whole files.

\begin{figure*}[t]
\centering
\includegraphics[width=0.9\linewidth]{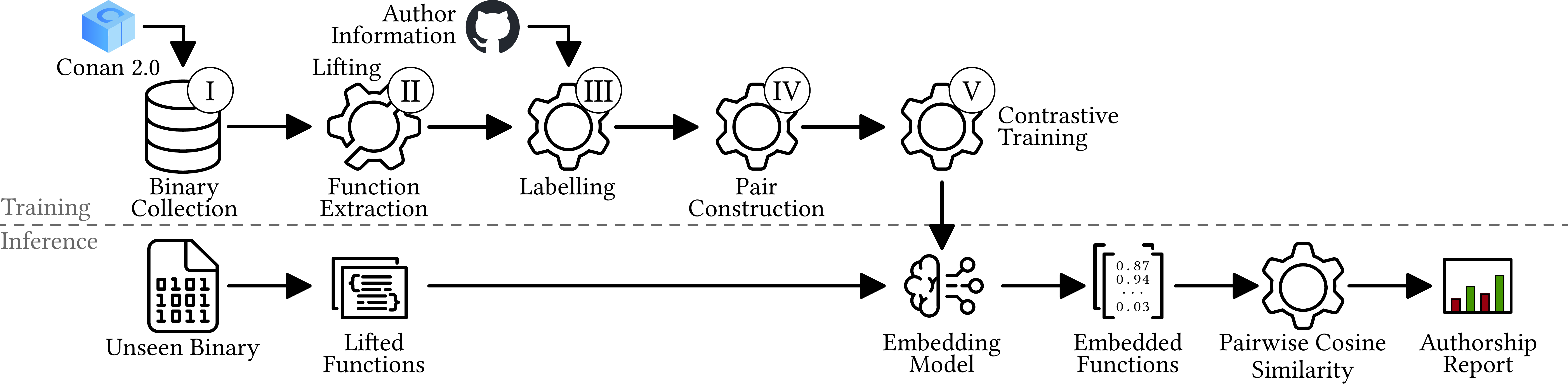}
    \caption{Training and inference of OCEAN. All steps are described in \Cref{sec:pipeline}.
    }
    \label{fig:system-description}
\end{figure*}

\subsection{Overview}
\label{sec:pipeline}
\noindent\textbf{Training.} We utilize \dataset{} for training (Step I), which consists of program binaries together with the source code authored by multiple programmers (see also \Cref{fig:system-description} for a full overview of the pipeline). In the representation extraction phase (Step II), we extract functions in the desired representation from each binary using Ghidra, a state-of-the-art reverse engineering tool~\cite{ghidra}.
Given a representation and the function-level authorship information of \dataset, we build positive and negative pairs for training (Step III). Positive pairs consist of samples from the same author, 
while negative pairs combine samples from different authors. Using these pairs, we train a contrastive learning model to create embeddings for the samples (Step IV). The model learns to map samples from the same author (positive pairs) closer in the embedding space while pushing samples from different authors (negative pairs) apart.

\noindent \textbf{Inference}. During the inference phase \schemename{} embeds samples pairwise into a high-dimensional vector space (Step V). To determine the authorship similarity, \schemename{} leverages the cosine similarity between the embeddings of the two samples (Step VI). If the similarity is above a predefined threshold $\theta$, \schemename{} classifies them to be authored by the same individual.

\subsection{Contrastive Learning}

Contrastive learning is a technique where the model learns to distinguish between similar and dissimilar pairs of data. Therefore, it is particularly useful for tasks that require the model to generate informative and discriminative embeddings. In our context, these embeddings are crucial, as the cosine distance between two embeddings is used to determine whether the two inputs leading to the embeddings are from the same author.

In \schemename{}, we use contrastive learning during training to ensure that the model can generate these rich embeddings. Specifically, we use a loss function that decreases the cosine similarity for negative pairs (different authors) and increases it for positive pairs (same author). Therefore, we adopt the loss function from \etal{Ai}~\cite{ai-etal-2022-whodunit}:
\[
\mathcal{L} = -\sum_i \log \left(\frac{\sum_{a_i=a_j} \exp \left(\cos{-}sim \left(e_i, e_j\right) / \tau\right)}{\sum_k \exp \left(\cos{-}sim \left(e_i, e_k\right) / \tau\right)}\right)
\]

Here, $e_i$ corresponds to the embedding generated by a model for the i-th sample, i.e., $e_i = f(x_i)$ and $\tau$ is a fixed constant for the temperature, e.g., $0.1$. The intuition behind the loss function is, that the numerator becomes large for positive samples ($a_i=a_j$), while the denominator adds a penalty for all negative samples that have a high cosine similarity.
Thus, the fraction gets closer to one as the classifier improves. The fraction is encapsulated in a logarithm to turn the multiplication of probabilities into a sum, which makes the loss easier to calculate. Thus, the sign is inverted with a minus, so that the smaller the loss, the closer the fraction is to one. 
This loss function ensures that negative samples have reduced cosine similarity values compared to positive samples.

After training, the model $f(x)$ can generate embeddings for any given sample. To compare two samples $x_1$ and $x_2$, we compute the cosine similarity between their embeddings as follows:
\[
c(x_1, x_2) = 1 \iff \frac{f(x_1) \cdot f(x_2)}{\|f(x_1)\| \|f(x_2)\|} > \theta
\]
As the model is trained to have a higher cosine similarity for samples from the same author, we infer that if the cosine similarity $c(x_1, x_2)$ exceeds a predefined threshold $\theta$, the samples are likely to be from the same author. Otherwise, they are assumed to be from different authors.

\subsection{\schemename{} for Security}
\label{sec:threshold-definition}

As \schemename{} can compare authors, it can detect when new authors have contributed to a dependency during a software update. This ability can be used to identify potential code injections in the software supply chain. We hypothesize that although many functions are modified or added during typical updates, a function injected by an attacker will exhibit stylistic discrepancies detectable by authorship identification techniques. This technique could be used, for example, to flag suspicious functions for further manual review by security analysts.

On every update, we can compare each changed or added function to all functions of the previous version.
To achieve this compare, we use our previously described pipeline (see \Cref{sec:pipeline}) to generate embedding vectors representing the stylistic features of each function in the old binary. Thus, we have a corpus of vectors for the old functions. For the updated binary, we collect all changed functions and compare each of these functions to all functions in the old corpus using the cosine distance. If the distance exceeds a certain threshold, we flag the new functions as possibly written by a new author.

However, given the multiple comparisons involved, adhering to a single predefined threshold for each individual comparison is not statistically robust. The high number of comparisons increase the likelihood of finding at least one function whose distance exceeds the threshold by chance. So, we need to calibrate the model~\cite{model_calibration} and dynamically adjust the threshold based on the number of changed functions to control false discovery rates.
To determine the threshold for detecting potentially malicious code, we use statistical models. Since our distances are between 0 and 1, we use a Beta distribution to approximate the similarities observed for samples from the same author. Given the Beta distribution, we can calculate the expected minimum of a multitude of comparisons. 
\begin{align*}
    f_{\text{min}}(x) &= n \cdot f(x) \cdot (1 - F(x))^{n-1}
\end{align*}

Where \( n \) is the number of comparisons, \( f(x) \) is the probability density function of the Beta distribution, and \( F(x) \) is its cumulative distribution function, both estimated from the data.
The expected minimum distance \( E(X_{\text{min}}) \) 
is calculated as:
\[
E(X_{\text{min}}) = \int_{0}^{1} x \cdot f_{\text{min}}(x) \, dx
\]
Hence, \( \theta =  E(X_{\text{min}}) \) enables us to establish a dynamic threshold that adapts based on the number of comparisons.
If a function has a bigger minimal distance to the corpus than $\theta$ it was probably modified or inserted by a new author. 

\subsection{Implementation}
Our implementation is based on ContraX~\cite{ai-etal-2022-whodunit}, a framework that uses contrastive learning to perform authorship identification on written text. We have extended ContraX to support several models from related work~\cite{bin_eye, Meng2017BinaryCM, omi2021multiauth, scsgan, DBLP:journals/access/ZafarSSM20, DBLP:conf/dimva/LooseMPB023} as well as some new ones.
These include Convolutional Neural Networks (CNN), Long Short-Term Memory (LSTM) networks, and Feedforward Neural Networks (FNN) using two different embedding styles: Term Frequency-Inverse Document Frequency (TF-IDF) and N-grams. In addition, we use a pre-trained BERT model, specifically UniXcoder~\cite{unixcoder}, which has been trained on code-related tasks to improve its code understanding. Notably, these tasks are unrelated to authorship identification. Research suggests that pre-trained models for code-related tasks often outperform models trained from scratch, due to their ability to leverage learned representations from a large corpus of code~\cite{Peculiar}.
While random forest or decision trees are often used in related work~\cite{10.5555/2831143.2831160, Dauber2017GitBW, multi_x, hao2022multiauth, DBLP:conf/sigsoft/BogomolovKRBB21}, we did not use those models in our approach because they do not naturally generate embeddings, which are necessary for contrastive learning.

Each model requires a custom pre-processing pipeline to fit within our framework. We ensured that every model received raw input, which was subsequently converted to the necessary format for the respective model. Therefore, all models share a common interface, simplifying the integration process.
For the LSTM, we implemented a custom Byte Pair Tokenizer, which was trained using the training dataset. 
For the FNN, we used two different configurations for preprocessing: The first configuration involved TF-IDF at the character level, while the second employed character-level n-grams ranging from $n=1$ to $n=3$. Both configurations were pre-trained using the training data and returned a maximum of 1000 features.
For the CNN, we followed the methodology outlined by \etal{Alrabaee}~\cite{bin_eye} and converted the bytes of the input data into a 100x100 matrix, effectively creating a 100x100 pixel grayscale image. If the input was too small, zeros were added for padding; if it was too large, the input was truncated. 
The hyperparameters of our models are shown in \Cref{tab:hyperparameters-for-all-models} in the Appendix. The output of each model is a 768-dimensional vector, rather than a classification value, to facilitate their integration into the contrastive learning framework. The cosine distance between two embedding vectors is then used for classification. 

\section{Experiments and Results}

To evaluate \schemename{} and benchmark it against the previous state-of-the-art solution in open world extreme authorship identification (SCS-Gan by \etal{Ou}~\cite{scsgan}), we examine several research questions. To the best of our knowledge \schemename{} is the only solution supporting compiled binaries in the extreme open world setting. Therefore, we perform the comparison with SCS-Gan using only source-code representations. We evaluate the performance of \schemename{} on binary representations in a more general approach without comparisons to previous work, as there seems to be no scheme in the same setting enabling a fair comparison. Additionally, we evaluate \dataset{} against the Google Code Jam (GCJ) dataset. After answering all research questions, we further demonstrate the capabilities of \schemename{} in two case studies. 

\subsection{Research Questions}
\begin{itemize}[noitemsep,topsep=0pt,leftmargin=*]

    \item \textbf{RQ1:} How does \schemename{} compare to SCS-Gan in terms of classification performance on source code representations?

    \item \textbf{RQ2:} Does \dataset{} enable the development of authorship identification systems with better real world performance than GCJ? 
    
    \item \textbf{RQ3:} What is the best combination of program representation and model in \schemename{} to perform extreme authorship identification on binary representations? 

    \item \textbf{RQ4:} How to leverage \schemename{} in a real world application?

    \item \textbf{RQ5:} How do Ghidra warnings affect our performance?

    \item \textbf{RQ6:} How robust is \schemename{} against compiler optimizations?
\end{itemize}

\subsection{Experiment Setup}
\label{sec:eval:experiment_setup}
For all experiments, unless otherwise noted, we use function level granularity and binaries compiled with \textit{O0} flags, i.e., no optimization. A function is attributed to an author if they contributed at least 51\% of it, as 51\% is the minimum threshold for the majority of the function to be written by that author.
Functions longer than 1000 tokens are truncated to fit the input requirements of the model~\cite{unixcoder}. 
This limit was chosen to provide support for long inputs while  also maintaining computational efficiency. 
If a function is shorter, it is padded to 1000 tokens.
Additionally, we always evaluate pairs from the same binary to make sure that we evaluate stylistic detection rather than binary detection.

\begin{table}[t]
    \setlength\tabcolsep{3.4pt}
    \centering
    \begin{threeparttable}
    \caption{Performance comparison of OCEAN and SCS-Gan~\cite{scsgan}). SCS-Gan results with an asterisk (*) are replicated by us, while those with a dagger (\dag) are from the original paper. }
    \begin{tabularx}{\columnwidth}{Xrrrrr}
    \toprule
         & AUROC & ACC & PRC & REC & F1 \\ 
    \midrule
        SCS-Gan (16 head)\dag    & 0.96 & 0.87 & 0.96 & 0.71 & 0.82 \\ 
        SCS-Gan (16 head+GAN)\dag & 0.96 & 0.90 & 0.97 & 0.78 & 0.86 \\ 
        SCS-Gan (16 head, 40 epochs)* & 0.91 & 0.92 & 0.77 & 0.63 & 0.69 \\ 
        SCS-Gan (16 head+GAN, 20 epochs)* & 0.85 & 0.88 & 0.55 & 0.55 & 0.55 \\ 
        \schemename{} & \textbf{0.99} & \textbf{0.95} & \textbf{0.98} & \textbf{0.92} & \textbf{0.95} \\ \bottomrule
    \end{tabularx}
    \label{tab:eval:compare_to_related_work}
    \end{threeparttable}
\end{table}

\noindent \textbf{\schemename{} vs SCS-Gan:} To answer \textbf{RQ1}, we compare \schemename{} to the state of the art in extreme author identification, namely SCS-Gan\cite{scsgan} by \etal{Ou} To perform this comparison, we initially contacted the authors to retrieve the source code, which they now published on GitHub\footnote{\url{https://github.com/L1NNA/SourceCodeAuthorshipAnalysis/tree/SCS-Gan}}. For a fair comparison, we used their code and followed their methodology by leveraging the 2017 subset of GCJ and working on file-level granularity using source code for this experiment.
Our reproduction of the experiments of SGS-Gan result in a degraded performance for SGS-Gan. We attribute the discrepancy to the following factors:
The provided code did not include data preprocessing, so it was necessary to independently implement data preprocessing, for which we utilized the same setup as for \schemename{}. Additionally, due to occasional failures in training the SCS-Gan classifier, we had to limit training to 20 epochs instead of the recommended 100.
Finally, we opted to use UniXcoder~\cite{unixcoder} instead of CodeBERT~\cite{codebert} for the GAN component of SCS-Gan. UniXcoder, created by the same team as CodeBert, is a successor model that supports \texttt{C} and \texttt{C++}, which should have a positive impact on performance.
We benchmark our results against both the original results reported by \etal{Ou} and our reproduction of SCS-Gan in our performance comparison. The results are shown in \Cref{tab:eval:compare_to_related_work}.
\schemename{} achieves the highest scores in all metrics, even if not compared to our replicated but the original results from \etal{Ou}, indicating its superior performance compared to existing methodologies. For performance metrics that depend on a threshold value, we selected $\theta = 0.84$ as a minimal value of the cosine similarity so that the pair is predicted as \textit{same author}. This threshold was chosen to optimize the F1 score on a different and unrelated dataset, i.e., \dataset{} and \snoopy{}.
 
\begin{answerbox}
\textbf{RQ1:} \schemename{} outperforms SCS-Gan by a large margin in the F1 score.
\end{answerbox}

\begin{table}[t]
    \setlength\tabcolsep{3.4pt}
    \centering
    \begin{threeparttable}
    \caption{RQ2: Comparison of model performance on test split and control sample (SNOOPY) when trained on GCJ and CONAN. When trained on CONAN, all models perform better on SNOOPY.}

\begin{tabularx}{\columnwidth}{XcXcc}
\toprule
      & \multicolumn{2}{l}{Trained on GCJ} & \multicolumn{2}{l}{Trained on Conan} \\ 
Model & Test & \snoopy & Test & \snoopy \\ \midrule
FNN TF 
            & $0.81$
            & $0.70$
            & $0.78$
            & \textbf{0.72} \\
             
FNN NG 
            & $0.89$
            & $0.68$
            & $0.82$
            & \textbf{0.76} \\
             
UniXcoder 
            & $0.95$
            & $0.88$
            & $0.96$
            & \textbf{0.94} \\
             
Average 
            & $0.88$
            & $0.75$
            & $0.86$
            & \textbf{0.81} \\
            \bottomrule
\end{tabularx}

    \label{tab:gcj_vs_our_dataset}
    \end{threeparttable}
\end{table}

\noindent \textbf{\dataset{} vs. GCJ:} To answer \textbf{RQ2}, we compare \dataset{} (cf.~\Cref{ssec:Conan}) to the GCJ
dataset, which is commonly used in related work \cite{bin_author, bin_eye,10.5555/2831143.2831160, multi_x, scsgan, DBLP:conf/icse/LiCCZX22, DBLP:journals/access/ZafarSSM20, DBLP:conf/sigsoft/BogomolovKRBB21}.
We trained all models that work in our contrastive learning setup (\textit{see RQ3}) on GCJ and \dataset{} respectively. The evaluation was conducted on the test split of the training datasets and \snoopy. As all samples in GCJ are written by a single author, we discarded all samples from \dataset{} and \snoopy{} that were written by more than one author to allow for an unbiased comparison.
As representation, we used the original source code, consistent with the predominant usage of the GCJ dataset in previous studies \cite{multi_x, 10.5555/2831143.2831160, DBLP:conf/icse/LiCCZX22, scsgan, DBLP:journals/access/ZafarSSM20, DBLP:conf/sigsoft/BogomolovKRBB21}.
We used an 80/20 split for training and evaluation with respect to the authors in the training datasets to ensure an appropriate split for training and validation purposes. By splitting by authors, we ensured that none of the authors used for evaluation were included in the training set. In addition, we run all experiments 3 times and report the average. As the median standard deviation was 0.01, we do not include it in the results table.

As shown in \Cref{tab:gcj_vs_our_dataset}, when inspecting the test split of the training data, models trained on the GCJ dataset generally performed better than those trained on \dataset{}. 
The superior performance on the training data could be attributed to the larger size and potentially more homogeneous nature of the GCJ dataset.
However, when evaluated on the unseen \snoopy{} dataset, all models performed better when trained on the \dataset{} dataset. Especially UniXcoder, which exhibited a 7\% performance improvement when trained on \dataset{}. Overall, UniXcoder outperformed all other models by a significant margin, achieving the best results with a performance gap of 18 percentage points compared to the next best performing model. These results suggest, that top performance can only be achieved using our \dataset{} dataset together with a strong model.

This comparison confirms the assertion by Abazari \emph{et al.} \cite{dataset} and \etal{Bogomolov} \cite{DBLP:conf/sigsoft/BogomolovKRBB21} that the GCJ dataset provides an overly optimistic evaluation for the training dataset, while its real-world performance is not reliable. Additionally, evaluating solely on the training dataset's test split risks data snooping \cite{DBLP:conf/uss/Arp/dos-and-donts}.

\begin{answerbox}
    \textbf{RQ2:} Using the \dataset{} dataset improves the performance on real-world codebases, indicating its superior value for real world applications compared to the GCJ dataset.
\end{answerbox}

\label{sec:eval:best_representation_and_model}
\noindent \textbf{Models vs. Representations:} In order to address \textbf{RQ3}, we performed a grid search over ML models and code representations. Specifically, we evaluated the performance of a pretrained Bert Model, namely UniXcoder \cite{unixcoder}, UniXcoder without pretraining, a CNN, a FNN, and an LSTM model on five different code representations: Assembly, Source Code, P-Code, the raw binary, and decompiled C code, including their cleaned variants (see \Cref{sec:background:program-representations}). These experiments were conducted using \dataset{} and evaluated against \snoopy{}.
Given those 11 representations and 6 different models result in a search space of 66 combinations. To ensure robustness, we repeated all experiments 3 times. A complete overview of all combinations is provided in \Cref{tab:model_vs_repr_full} in the Appendix. Both show the AUC score, as this metric does not require a self-defined threshold $\theta$.

As shown in \Cref{tab:model_vs_repr_full}, all results are stable, with an average standard deviation of 0.02. However, several models such as CNN, LSTM, and UniXcoder without pre-training (\textit{UniXcoder (F)}), do not perform well in our contrastive learning setup. These results are consistent with previous observations indicating that CNN and LSTM models often require extensive preprocessing to perform effectively \cite{bin_eye, Meng2017BinaryCM, omi2021multiauth, scsgan}. UniXcoder without pre-training performs significantly worse than UniXcoder with pre-training. This finding is consistent with the work of Wu et al. \cite{Peculiar}, who emphasize the critical role of pre-training in improving model performance.
The best performance have UniXcoder with pre-trained weights and the FNN. The UniXcoder with pre-trained weights achieved an AUC of up to 0.92, significantly outperforming the same model without pre-training (max AUC of 0.80). 
However, the FNN also performs reasonably well as a smaller model without pre-training. Both the TF-IDF and N-gram representations have an average AUC of 0.68 and 0.71 across all representations. While they have a similar maximum performance as UniXcoder without pre-training, they are more stable across all representations. We speculate that the stability stems from the simpler network structure with fewer weights. Therefore, the model can use all representations up to a certain point, beyond which nuances become important. 

Regarding the representations, it is unsurprising that source code performs best (average AUROC score of 0.73 across all models). This representation is written directly by a human, preserving all stylistic authorship information. However, whether or not comments are included makes no difference, indicating that the code is sufficient for authorship identification. Even decompiled code works effectively with an average AUROC score of 0.71, especially for UniXcoder (AUROC score of 0.84), which aligns with its pre-training data. However, the decompiler sometimes inserts comments during decompilation (see RQ5). While the version where all inserted comments removed performs slightly better (AUROC score of 0.86), the difference is small.

Interestingly, the raw assembly and the cleaned assembly representation show a significant performance difference of 9 percent points. The removal of hex values results in a significant performance difference because they also correspond to variable usage information. The hex values help the model to understand the flow and structure of the code. However, the raw assembly version performs very similarly to decompiled code with an AUROC score of 0.85. It appears as if the model inferred higher-level structures and patterns from the raw assembly, similar to the process of decompilation. Therefore, raw assembly and decompiled code seem to be equivalent for a strong model. If speed is important, one should choose assembly because it saves the step of decompilation. However, if time is not a critical factor, decompiled code is presumably the better representation because it has better values on average and UniXcoder is pre-trained on code as representation.

\begin{answerbox}
\textbf{RQ3:} The best representation for authorship identification on binaries is decompiled code without comments together with a strong model.
\end{answerbox}

\begin{figure}[t]
    \centering
    \includegraphics[width=\columnwidth]{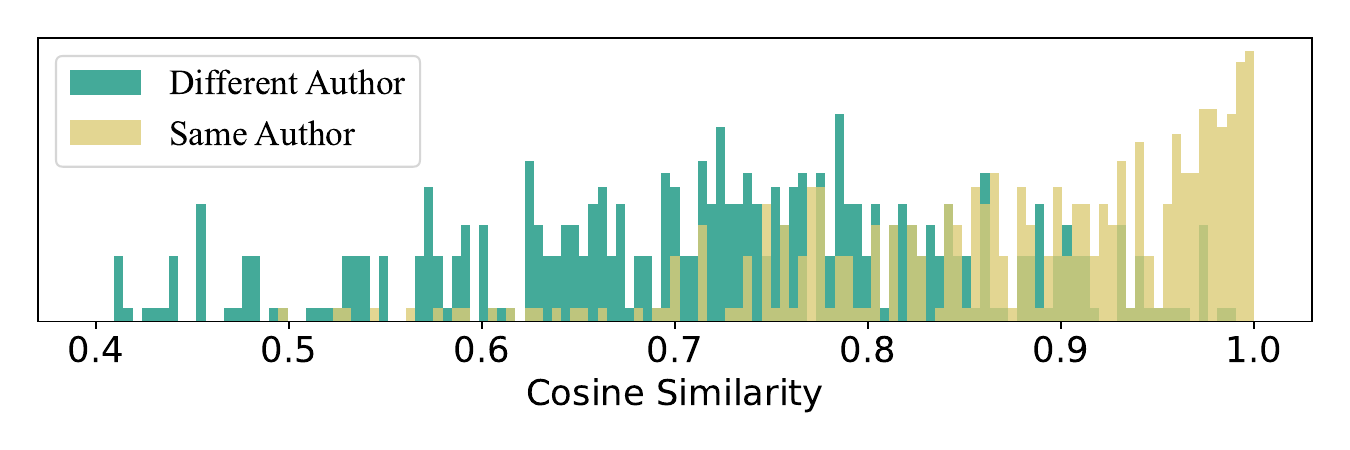}
    \caption{RQ4: Histogram of cosine similarities of an unseen dataset for the same and different authors. The expected minimum for same authors is at $\theta = 0.72$. }
    \label{fig:eval:CS:inter-vs-intra}
\end{figure}

\noindent \textbf{Threshold determination:} We have evaluated \schemename{} using AUROC, a metric that does not require a specific threshold. However, in real-world applications, the system should output a clear verdict. Hence, \textbf{RQ4} investigates the threshold to determine if the author is the same or different.
We trained the classification model on the \dataset{} dataset and used \snoopy{} as an unseen dataset to generate a pair of functions from the same author and a pair of functions from different authors. Using the trained model, we generated embeddings for all these pairs and computed the cosine similarities.
\Cref{fig:eval:CS:inter-vs-intra} presents a histogram of the cosine similarities of those pairs. 
The median cosine similarity for different authors is 0.73, while the median cosine similarity for the same authors is 0.94.

To determine the optimal threshold $\theta$, we can use the formula from \Cref{sec:threshold-definition} together with the data from \Cref{fig:eval:CS:inter-vs-intra} and $n=1$ as we only compare one pair. This calculation results in the threshold $\theta = 0.72$.
This threshold is the expected minimum for same authors and defines when an author is considered different.

\begin{answerbox}

    \textbf{RQ4:} \schemename{} is applicable in real-world scenarios because it allows the use of a threshold to make unambiguous verdicts based on two observable distributions. 

\end{answerbox}

\subsubsection*{Ghidra performance}

\begin{figure}[t]
    \centering
    \includegraphics[width=\columnwidth]{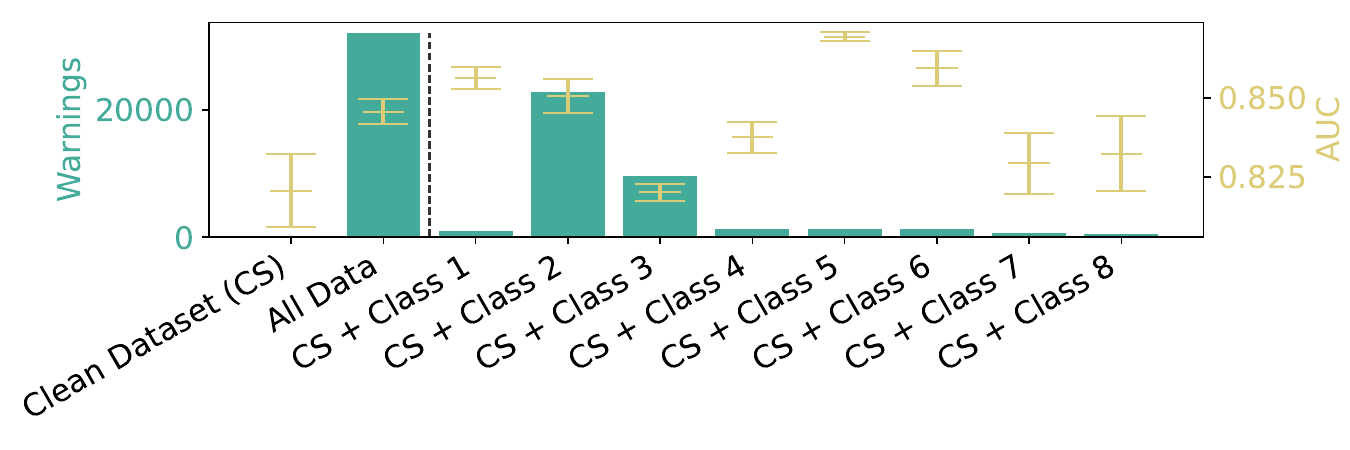}
    \caption{Performance comparison of samples with Ghidra warnings. Using samples with warnings does not decrease detection performance, but may increase it.}
    \label{fig:eval:error_classes}
\end{figure}

As Ghidra is the core of our data generation, the performance of our model is inherently tied to Ghidra's functionality. If Ghidra encounters certain errors, such as parsing failures or incorrect disassembly, our entire data generation process is compromised. Therefore, \textbf{RQ5} evaluates how warnings within Ghidra affect our performance.
During dataset generation, we identified 18 classes of warnings, e.g., \texttt{Removing unreachable block}. A comprehensive list of these warnings is given in \Cref{tab:appendix:warnings} in the Appendix. These error classes are not mutually exclusive, i.e., a single sample may contain multiple warnings. 
As experiment, we analyze all warning classes that appear in more than 1\% of the data. As two baselines, we define \textit{Clean Dataset (CS)} as the dataset without any samples having warnings and \textit{All Data} including all samples regardless of warnings. We then evaluate performance using CS combined with each warning class individually. As always, we use the \dataset{} dataset for training and \snoopy{} for evaluation, but this time only the specified subset, depending on the allowed warning type. The results are shown in \Cref{fig:eval:error_classes}.

First, it can be seen, the amount of data varies significantly depending on whether data containing warnings is included. For example, warning class 1 represents 0.51\% of the dataset. Despite these variations, the overall performance impact of including or not including warnings is small. The dataset with no warnings (CS) yields an AUC of 0.82, while including all warnings yields an AUC of 0.85. Additionally, the size of the warning class is not related to the AUC score, suggesting that the differences in performance are due to the data and not to different training dataset sizes. 
Interestingly, some classes, such as warning class 5, \textit{Bad instruction - Truncating control flow here}, increase performance when included, achieving an AUC score of 0.87. So in general, the dataset with no warnings performs the worst, suggesting that certain coding practices that trigger warnings may make programmers better identifiable.

\begin{answerbox}
    \textbf{RQ5:} Samples that trigger warnings in Ghidra do not affect the performance of \schemename{}. In contrast, some slightly increase the performance of the model.
\end{answerbox}

\noindent \textbf{Robustness:} 
So far, we have only evaluated binaries compiled with \textit{O0} flags, which lack compiler optimization and are therefore larger and slower. Such binaries are rarely used in practice, as developers typically use compiler flags for optimization. Therefore, \textbf{RQ6} investigates whether optimization affects our performance.
For this experiment, we use the \dataset{} dataset, which uses a set of debugging compilation flags in supported projects and the original compilation flags for the remaining dependencies, thus providing a wide variety of compilation settings. This diversity makes it suitable as a training dataset. We also recompiled \snoopy{} with all optimization levels from \textit{O0} (no optimization) to \textit{O3} (high optimization) to measure the performance across different optimization levels. However, due to compilation issues with PHP and GLibC under certain settings, we do not include them in this evaluation. Therefore, only a subset of \snoopy{} was used for this experiment.

Using the \dataset{} dataset, we trained a model and evaluated it on \snoopy{} at different optimization levels. The results show that the performance at the \textit{O2} optimization level is the highest, with an AUC of 0.86, while the performance at \textit{O0} and \textit{O3} is slightly lower at 0.82. The performance for \textit{O1} is 0.83. Thus, the performance of our method is relatively consistent across different optimization levels, with slight variations. 
To ensure that these results were not just due to Ghidra's effectiveness in decompiling \textit{O1} and \textit{O2} close to the original representation, we repeated the experiment using raw bytes as input. Similar correlations were observed, with \textit{O1} and \textit{O2} again performing best. A detailed comparison is presented in the Appendix in \Cref{tab:appendix:optimization_level}.
There could be two possible reasons for the high performance at \textit{O2}. First, \textit{O2} may be the most common level of optimization in the \dataset{} dataset. Second, as noted by Meng \emph{et al.} \cite{Meng2017BinaryCM}, compiler optimizations can remove features introduced by the compiler itself that are more distracting than useful for author identification. Therefore, higher optimization levels potentially improve performance. However, we cannot confirm \etal{Meng} findings that \textit{O3} improves performance even further, because the heavy function inlining potentially damages some ground truth labels on function granularity when merging multiple functions from multiple authors into one.

\begin{answerbox}
    \textbf{RQ6}: Compiler optimizations do not degrade the performance of \schemename{}, supporting its applicability in real-world scenarios where optimized binaries are prevalent.  
\end{answerbox}

\begin{figure}[t]
    \centering
    \includegraphics[width=\columnwidth]{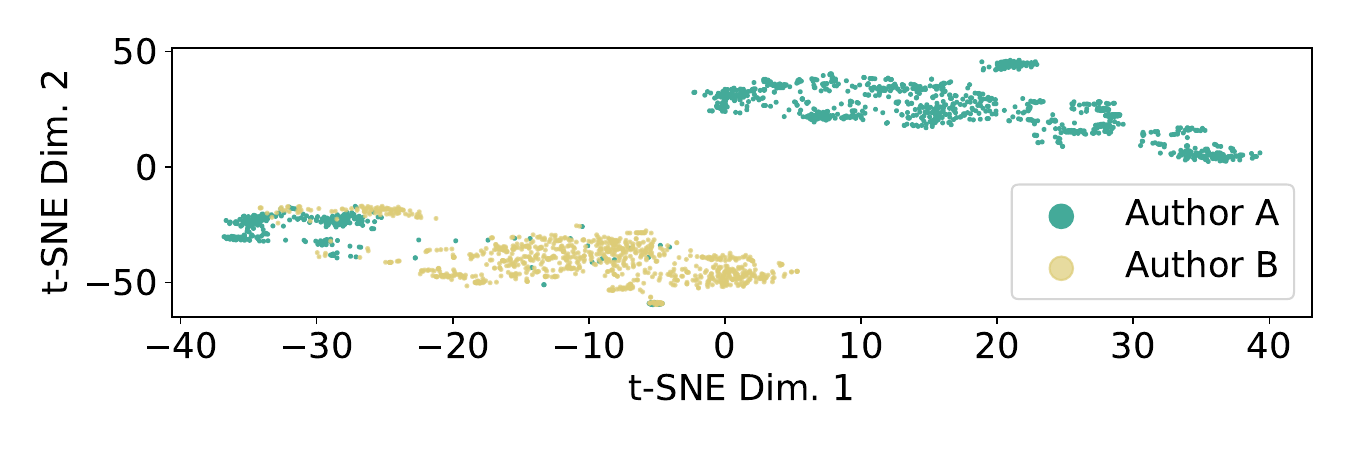}
    \caption{CS1: t-SNE visualization of embedding vectors for functions written by two authors. Each dot represents a function, with colors indicating the author.} 

    \label{fig:eval:CS:tsne}
\end{figure}

\subsection{Case Study}

Having assessed the overall performance of our authorship identification system, we now demonstrate its practical applications and capabilities by exploring two case studies.

\begin{itemize}
    \item \textbf{CS1:} Can \schemename{} distinguish two different authors in a compiled and unseen application?

    \item \textbf{CS2:} Can \schemename{} be used to identify a new author between two software versions?
\end{itemize}

\noindent \textbf{Visualization of two authors:} To address \textbf{CS1}, we compare the embeddings vectors of functions written by two authors. The expectation is that the vectors for each author will form a cluster. As target, we choose PHP because it has the highest number of functions written by two authors, making the case study more representative.
For both authors, we use all their samples, i.e., 1456 for Author A and 986 for Author B. We then create embeddings using a model that has been trained on decompiled and cleaned code from the \dataset{} dataset. For every resulting embedding vector, we calculate the cosine distance to all other vectors. Using t-Distributed Stochastic Neighbor Embedding (t-SNE) we visually represent these relationships in \Cref{fig:eval:CS:tsne}. 

It can be seen that there are indeed two distinct clusters, with each dot representing a single function written by one of the two authors. While the majority of dots clearly belong to one of the two clusters, there are a few dots that appear in the opposite cluster. In particular, in the upper left of Author B's cluster, there is a small sub-cluster of dots written by Author A. 
However, the clear separation of the majority of points indicates that \schemename{} can effectively distinguish between authors based on their coding patterns.

\begin{answerbox}

    \textbf{CS1:} \schemename{} is effective in distinguishing between authors based on their coding patterns, as evidenced by the formation of two clusters.
\end{answerbox}

\begin{figure}[t]
    \centering
    \includegraphics[width=1\columnwidth]{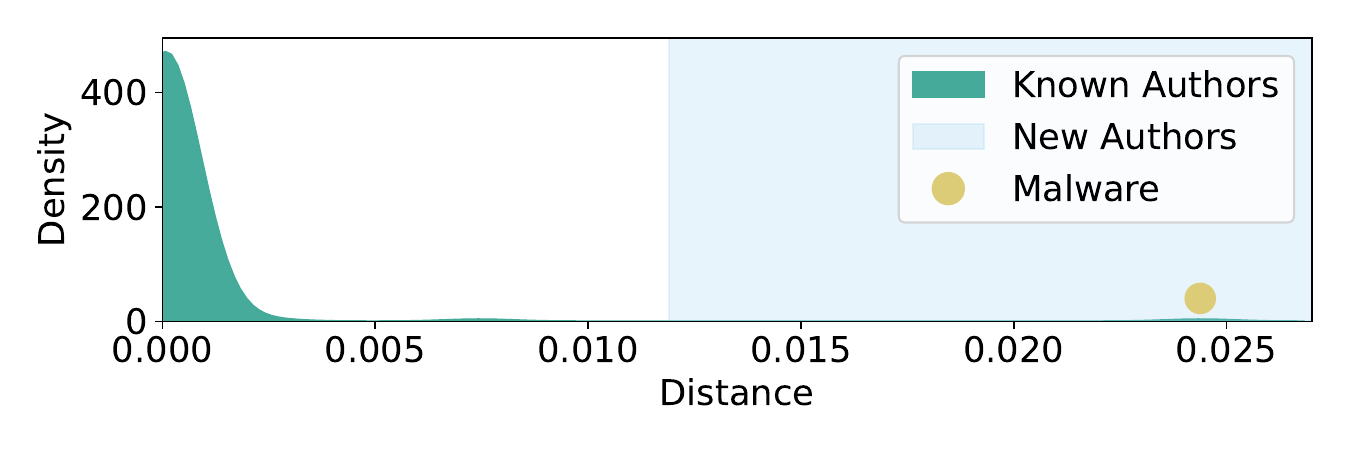}
    \caption{
    Approximation of the probability distribution of distances for known authors. The malware exceeds this threshold, making it detectable by our method.}
    \label{fig:eval:malware-plot}
    
\end{figure}
\noindent \textbf{Security:} 
To answer \textbf{CS2}, we test the feasibility of detecting a new author between two software versions. Therefore, we simulate a compromised software supply chain, by introducing malware in a software update. 
We used the malicious code from \etal{Zhou}~\cite{DBLP:conf/ccs/ZhouGJEJ18/injected_malware} 
and merged it into a single function. The malicious sample encrypts the entire file system, a common approach for malware often referred to as \emph{ransomware}. Next, we transformed the code from using Windows-based encryption libraries to using Linux-based equivalents. To minimize our influence on the injected code, we used ChatGPT4-omni~\cite{gpt4} to rewrite the code and removed all comments that were not in the original malware sample. After injecting the malware into the target software, we used the style check tools from the target project to match the style and coding guidelines used in the project.
Specifically, we injected the malware into OpenSSH subset of \snoopy, 
between version v9.7 and v9.8. So v9.7 is unmodified while v9.8 contains the malware by \etal{Zhou} We chose OpenSSH due to its widespread use and critical role in secure communications. Using Ghidra, we decompiled both versions and automatically identified functions that changed or have been added.
Version v9.7 contains 1293 functions in total. During the update, 406 functions were changed or added, one of which was the injected malware. 
We treated all functions that contained at least 50 assembly instructions as potentially malicious, as OpenSSH includes many functions with very little statements after decompilation in Ghidra. This left us with 137 potentially malicious functions.
Each of these functions was compared to the 1293 functions in the old version of the binary, and we measured the minimum distance for each comparison.
Using the analysis from \Cref{sec:threshold-definition} with the data from \Cref{fig:eval:CS:inter-vs-intra} and $n=1293$, we get the threshold $\theta = 0.0119$. Hence, we treat every distance greater than $0.0119$ as potentially malicious, as shown in \Cref{fig:eval:malware-plot}. Our injected malware clearly stood out as malicious, by being the only function above the defined threshold.

\begin{answerbox}
    \textbf{CS2:} \schemename{} effectively detects a new author between software versions, thus helping to identify a compromised software supply chain.
\end{answerbox}

\section{Related Work}
\label{sec:rel_work}

Authorship identification has long intrigued scholars, famously exemplified by the debate over Shakespeare's true identity. Leveraging stylometry, researchers have successfully attributed authorship in written texts\cite{sari-etal-2018-topic, ai-etal-2022-whodunit, AA_text_survey}. More recently, similar methods have been proposed for source code. 

Alrabaee \emph{et al.} \cite{bin_author, bin_eye} developed a pipeline that elevates program binaries into a canonical form and creates features from both the original binary and the canonical form.
Having a large database of compiler and open-source libraries, they detected general design choices made by the developers. Using these design choices, they attributed the input to a single known author or determined that it is an unknown author. Thus, they operated in the semi-open world scenario.
Meng \emph{et al.} \cite{meng_mult_auth} and Omi \emph{et al.} \cite{omi2021multiauth} both considered multiple authors per input file, but only in the closed world case.
\etal{Meng} chose an approach at the binary level by identifying neighboring basic blocks (a sequence of instructions without diverging control flow) by a single author and merging these blocks into larger single-author snippets.
They then analyzed those snippets regarding operand distribution, control flow, data flow and register usage.
Finally, they used a random forest classifier to predict the contributing authors. \etal{Omi} chose a different approach by converting the binary into an abstract syntax tree and training a deep neural network based on features extracted from the tree.
Both approaches were evaluated on relatively small datasets, with \etal{Meng} using 284 authors and \etal{Omi} using only 40 authors.
\etal{Bogomolov} \cite{DBLP:conf/sigsoft/BogomolovKRBB21} also considered multiple authors per file in the closed world scenario. They used a Random Forest and a Neural Network to distinguish between functions that have been written by a single author in a multi-author project. In their work, they highlighted the importance of a realistic dataset and identified several criteria for realistic datasets, such as a long project period and different work contexts with different code functionalities. Although they only considered freshly added functions to the code base in a closed world scenario, they showed that previous work already struggled in those slightly more complex scenarios. However, their approach used a closed world scenario and relied on large numbers of samples per author.

Caliskan \emph{et al.}~\cite{10.5555/2831143.2831160} explored a semi-open world, two-class scenario using a one-vs-all training approach by training models with data from one author as a single label and multiple authors as a combined label. During inference, the model predicted whether a sample was from the known author. Therefore, this approach fails in the extreme authorship attribution scenario.
The solutions of \etal{Dauber} \cite{Dauber2017GitBW} and \etal{Zafar} \cite{DBLP:journals/access/ZafarSSM20} showed whether an author does not belong to a set of suspected authors using the confidence of a classifier and a threshold. Hence, they were also in the semi-open world scenario. 
Additionally, all the above approaches required multiple samples per author in the training dataset, limiting them to a closed or semi-open world scenario.
We consider a much more challenging and less explored scenario in which we only need two samples to identify whether they are from the same author.

Abuhamad \emph{et al.} \cite{multi_x} proposed a classifier that operates at line-level granularity. They preprocessed source code by creating pairs of lines and trained a recurrent neural network (RNN) to classify if two lines were from the same author. They iteratively joined lines until blocks were formed, which were then used for the final author prediction. For the final prediction, they used another RNN that was trained on tokenized sequences gained from the merged sample files.
Hao et al. \cite{hao2022multiauth} improved on the approach of Abuhamad \emph{et al.} by extracting path-based features from each line of code before merging lines of code using a Siamese RNN. A Siamese Network consists of two identical neural networks that process input pairs and learn to distinguish between them. In the end they used a Random Forest classifier for the final author prediction. However, both approaches have the same problem for real-world applicability: They can not consider compiled binaries because the decompiler could mix the style of multiple lines. Mixing lines would cause any approach with line-granularity to fail because the classifier would merge these lines into incorrect blocks. We therefore need an approach that is more robust to noise over neighboring lines.

Ou \emph{et al.} \cite{scsgan} were the first to address the extreme authorship identification problem, a true open world scenario, by using a Siamese Network trained to predict whether two samples are from the same author. They used a generative adversarial network (GAN) in conjunction with a pretrained model focused on code semantics. The GAN was trained to generate realistic code samples, while the pretrained model helped in distinguishing stylistic features from semantic content. 
However, they evaluated their methods on the source code of GCJ, which in previous work has been shown to be a too optimistic dataset for real-world scenario \cite{dataset}. Additionally, they used a file granularity, assuming that only a single author contributed to a source code file. In contrast, our approach employs real-world code from GitHub in compiled format at the function granularity, thus dropping the single-author assumption per program and providing a more practical solution.

\section{Conclusion}

In this paper, we presented \schemename{}, a novel approach to authorship attribution of binary files using contrastive learning. By focusing on extreme authorship identification in an open-world scenario, our system goes beyond the traditional limitations of closed-world assumptions, providing a robust solution for real-world applications where new authors may emerge and source code may be unavailable.
In addition, we have introduced two novel datasets, \dataset{} and \snoopy{}, which better reflect real-world scenarios compared to previously used datasets such as Google Code Jam. Our extensive evaluations show that \schemename{} not only significantly outperforms the previous state-of-the-art, SCS-Gan, in the source code scenario, but also achieves impressive performance on compiled binaries, i.e., practical real-world data samples.
OCEAN can also detect code from new or unexpected authors, showing its potential for protecting the software supply chain. 
For now, our approach  attributes code to single authors, which necessitates the exclusion of multi-author samples. Addressing this limitation should be a focus of future research, and our framework, along with the datasets provided, is a strong foundation for further exploration in this area.

\bibliographystyle{plain}
\bibliography{
    References.bib
}
\newpage
\onecolumn
\appendixstyle
\appendix
\section{Appendix}
\noindent
\begin{table*}[ht]
\centering

\scalebox{1}{
\begin{tabular}{lcccccc|c}
\hline
Representation & CNN & FNN NG & FNN TF & LSTM & UniXcoder & UniXc. (F) & Average \\
\hline
Assembly (C) & $0.57 \pm 0.02$ & $0.71 \pm 0.05$ & $0.67 \pm 0.06$ & $0.54 \pm 0.02$ & $0.76 \pm 0.00$ & $0.66 \pm 0.00$ & $0.65 \pm 0.02$ \\
Assembly & $0.55 \pm 0.03$ & $0.73 \pm 0.07$ & $0.67 \pm 0.03$ & $0.52 \pm 0.04$ & $0.85 \pm 0.03$ & $0.59 \pm 0.00$ & $0.65 \pm 0.03$ \\
Decompiled & $0.60 \pm 0.05$ & $0.77 \pm 0.05$ & $0.76 \pm 0.03$ & $0.52 \pm 0.04$ & $0.84 \pm 0.00$ & $0.76 \pm 0.00$ & $0.71 \pm 0.03$ \\
Decompiled (C) & $0.59 \pm 0.03$ & $0.78 \pm 0.04$ & $0.74 \pm 0.05$ & $0.51 \pm 0.03$ & $0.86 \pm 0.00$ & $0.76 \pm 0.00$ & $0.70 \pm 0.03$ \\
P-Code & $0.65 \pm 0.04$ & $0.65 \pm 0.04$ & $0.64 \pm 0.04$ & $0.52 \pm 0.03$ & $0.74 \pm 0.00$ & $0.50 \pm 0.00$ & $0.61 \pm 0.02$ \\
P-Code (C) & $0.63 \pm 0.03$ & $0.64 \pm 0.03$ & $0.64 \pm 0.02$ & $0.52 \pm 0.02$ & $0.69 \pm 0.00$ & $0.52 \pm 0.00$ & $0.61 \pm 0.02$ \\
P-Code 2 & $0.56 \pm 0.03$ & $0.60 \pm 0.06$ & $0.60 \pm 0.04$ & $0.61 \pm 0.02$ & $0.79 \pm 0.00$ & $0.53 \pm 0.00$ & $0.61 \pm 0.02$ \\
P-Code 2 (C) & $0.55 \pm 0.02$ & $0.59 \pm 0.04$ & $0.59 \pm 0.05$ & $0.54 \pm 0.02$ & $0.63 \pm 0.00$ & $0.50 \pm 0.00$ & $0.57 \pm 0.02$ \\
Raw & $0.57 \pm 0.01$ & $0.70 \pm 0.02$ & $0.62 \pm 0.02$ & $0.53 \pm 0.00$ & $0.75 \pm 0.00$ & $0.64 \pm 0.00$ & $0.63 \pm 0.01$ \\
Source Code & $0.59 \pm 0.01$ & $0.79 \pm 0.06$ & $0.75 \pm 0.06$ & $0.54 \pm 0.04$ & $0.92 \pm 0.00$ & $0.80 \pm 0.00$ & $0.73 \pm 0.03$ \\
Source Code (C) & $0.59 \pm 0.03$ & $0.81 \pm 0.05$ & $0.77 \pm 0.05$ & $0.50 \pm 0.00$ & $0.92 \pm 0.00$ & $0.77 \pm 0.00$ & $0.73 \pm 0.02$ \\ \hline
Average & $0.59 \pm 0.03$ & $0.71 \pm 0.05$ & $0.68 \pm 0.04$ & $0.52 \pm 0.02$ & $0.80 \pm 0.00$ & $0.64 \pm 0.00$ & $0.66 \pm 0.02$ \\
\end{tabular}

} 

\caption{Average AUROC scores for all models and representations}
\label{tab:model_vs_repr_full}
\end{table*}

\noindent
\begin{table}[ht]
    \centering
    \begin{tabular}{lrrrr}
       Representation  & \textit{O0} & \textit{O1} & \textit{O2} & \textit{O3} \\ \hline
       Decompiled (C)  & 0.82 & 0.83 & 0.86 & 0.82 \\
       Raw & 0.65 & 0.75 & 0.76 & 0.74
    \end{tabular}
    \caption{Results for different levels of compiler optimization.}
    \label{tab:appendix:optimization_level}
\end{table}
\noindent
\begin{table}[ht]
    \centering
    \caption{Hyperparameters for all models used}
    \label{tab:hyperparameters-for-all-models}
    \subfloat[LSTM Model]{
        \begin{tabular}{l|l}
            \textbf{Parameter} & \textbf{Value} \\
            \hline
            vocab\_size & 1000 \\
            embed\_dim & 100 \\
            hidden\_dim & 64 \\
            output\_dim & 768 \\
            num\_lstm\_layers & 2 \\
        \end{tabular}
    }
    \quad
    \subfloat[FNN Model]{
        \begin{tabular}{l|l}
            \textbf{Parameter} & \textbf{Value} \\
            \hline
            input\_dim & 128 \\
            hidden\_dim & 64 \\
            output\_dim & 768 \\
            num\_hidden\_layer & 1 \\
        \end{tabular}
    } 
    \quad 
    \subfloat[CNN Model]{
        \begin{tabular}{l|l}
            \textbf{Parameter} & \textbf{Value} \\
            \hline
            input\_dim & $100^2$ \\
            conv\_kernel\_size & 2 \\
            conv\_stride & 2 \\
            out\_channels\_layer\_1 & 32 \\
            out\_channels\_layer\_2 & 128 \\
            out\_channels\_fc & 64 \\
            output\_dim & 768 \\
        \end{tabular}
    }
    \quad 
    \subfloat[UniXcoder]{
    \begin{tabular}{l}
        All parameters are as specified\\
     in the original paper\cite{unixcoder}.
    \end{tabular}
    }
\end{table}

\noindent

\begin{table*}[ht]
    \centering
    \scalebox{1}{
    \begin{tabular}{llrr}
    & \textbf{Warning} & \textbf{Count} & \textbf{Percentage} \\ \hline
    
\#1 & Removing unreachable block (ram,0x00000) & 38922 & 0.51 \\
\#2 & Subroutine does not return & 32694 & 0.43 \\
\#3 & Unknown calling convention & 9266 & 0.12 \\
\#4 & Load size is inaccurate & 2054 & 0.03 \\
\#5 & Bad instruction - Truncating control flow here & 1257 & 0.02 \\
\#6 & Control flow encountered bad instruction data & 1245 & 0.02 \\
\#7 & Globals starting with '\_' overlap smaller symbols at the same address & 596 & 0.01 \\
\#8 & Restarted to delay deadcode elimination for space: stack & 544 & 0.01 \\
\#9 & Unknown calling convention -- yet parameter storage is locked & 364 & 0.00 \\
\#10 & Ignoring partial resolution of indirect & 150 & 0.00 \\
\#11 & Could not recover jumptable at 0x00000. Too many branches & 114 & 0.00 \\
\#12 & Treating indirect jump as call & 114 & 0.00 \\
\#13 & Heritage AFTER dead removal. Example location: s0x00000 : 0x00000 & 106 & 0.00 \\
\#14 & Type propagation algorithm not settling & 77 & 0.00 \\
\#15 & Do nothing block with infinite loop & 3 & 0.00 \\
\#16 & Could not find normalized switch variable to match jumptable & 2 & 0.00 \\
\#17 & Restarted to delay deadcode elimination for space: ram & 2 & 0.00 \\
\#18 & Store size is inaccurate & 1 & 0.00 \\

\end{tabular}}

    \caption{Number of Ghidra-warnings for \dataset{} and \snoopy{}. }
    \label{tab:appendix:warnings}
\end{table*}

\clearpage
\subsection{Code Representations}
\label{appendix:code-representation-examples}
Section \ref{sec:background:program-representations} provides an overview of the various code representations used in this work. To illustrate these representations more concretely, the following examples showcase the function \texttt{match.c/match\_usergroup\_pattern\_list} from OpenSSH~\cite{snoopy-openssh}:

\vspace{.5cm}
\textbf{Source Code:}
\begin{lstlisting}[]
match_usergroup_pattern_list(const char *string, const char *pattern)
{
    #ifdef HAVE_CYGWIN
        /* Windows usernames may be Unicode and are not case sensitive */
        return cygwin_ug_match_pattern_list(string, pattern);
    #else
        /* Case sensitive match */
        return match_pattern_list(string, pattern, 0);
    #endif
}
\end{lstlisting}

\noindent\textbf{Source Code (cleaned):}
\begin{lstlisting}[]
match_usergroup_pattern_list(const char *string, const char *pattern)
{
    #ifdef HAVE_CYGWIN
        return cygwin_ug_match_pattern_list(string, pattern);
    #else
        return match_pattern_list(string, pattern, 0);
    #endif
}
\end{lstlisting}

\noindent\textbf{Decompiled:}
\begin{lstlisting}[]
int match_usergroup_pattern_list(char *string,char *pattern)
{
  int iVar1;
  char *pattern_local;
  char *string_local;
  iVar1 = match_pattern_list(string,pattern,0);
  return iVar1;
}
\end{lstlisting}

\noindent\begin{minipage}[t]{0.48\textwidth} 
\noindent\textbf{Assembly:}
\begin{lstlisting}
ENDBR64
PUSH RBP
MOV RBP,RSP
SUB RSP,0x10
MOV qword ptr [RBP + -0x8],RDI
MOV qword ptr [RBP + -0x10],RSI
MOV RCX,qword ptr [RBP + -0x10]
MOV RAX,qword ptr [RBP + -0x8]
MOV EDX,0x0
MOV RSI,RCX
MOV RDI,RAX
CALL 0x0017e20a
LEAVE
XOR EDX,EDX
XOR ECX,ECX
XOR ESI,ESI
XOR EDI,EDI
RET
\end{lstlisting}
\end{minipage}
\hfill 
\begin{minipage}[t]{0.48\textwidth}
\noindent\textbf{Assembly (cleaned):}
\begin{lstlisting}
ENDBR64
PUSH RBP
MOV RBP,RSP
SUB RSP,HEXSTR
MOV qword ptr [RBP + -HEXSTR],RDI
MOV qword ptr [RBP + -HEXSTR],RSI
MOV RCX,qword ptr [RBP + -HEXSTR]
MOV RAX,qword ptr [RBP + -HEXSTR]
MOV EDX,HEXSTR
MOV RSI,RCX
MOV RDI,RAX
CALL HEXSTR
LEAVE
XOR EDX,EDX
XOR ECX,ECX
XOR ESI,ESI
XOR EDI,EDI
RET
\end{lstlisting}
\end{minipage}

\noindent\begin{minipage}[t]{0.48\textwidth} 
\noindent\textbf{P-Code:}
\begin{lstlisting}[]
(unique, 0xf000, 8) COPY (register, 0x28, 8)
(register, 0x20, 8) INT_SUB (register, 0x20, 8)

(const, 0x8, 8)
STORE (const, 0x1b1, 8)
...
RETURN (register, 0x288, 8)
\end{lstlisting}  
\end{minipage}
\hfill 
\begin{minipage}[t]{0.48\textwidth}
\noindent\textbf{P-Code (cleaned):}
\begin{lstlisting}
(unique, HEXSTR, 8) COPY (register, HEXSTR, 8)
(register, HEXSTR, 8) INT_SUB (register, HEXSTR, 8)

(const, HEXSTR, 8)
STORE (const, HEXSTR, 8)
...
RETURN (register, HEXSTR, 8)
\end{lstlisting}
\end{minipage}

\noindent\begin{minipage}[t]{0.48\textwidth} 
\noindent\textbf{P-Code (2):}
\begin{lstlisting}[]
(register, 0x0, 4) CALL (ram, 0x17e20a, 8)
(register, 0x38, 8)
(register, 0x30, 8)
(const, 0x0, 4)
RETURN (const, 0x0, 8)
(register, 0x0, 4)
(register, 0x0, 4) COPY (register, 0x0, 4)
\end{lstlisting}  
\end{minipage}
\hfill 
\begin{minipage}[t]{0.48\textwidth}
\noindent\textbf{P-Code (2) (cleaned):}
\begin{lstlisting}
(register, HEXSTR, 4) CALL (ram, HEXSTR, 8)
(register, HEXSTR, 8)
(register, HEXSTR, 8)
(const, HEXSTR, 4)
RETURN (const, HEXSTR, 8)
(register, HEXSTR, 4)
(register, HEXSTR, 4) COPY (register, HEXSTR, 4)
\end{lstlisting}
\end{minipage}

\noindent\textbf{Raw bytes in hexadecimal:}
\begin{lstlisting}[]
f30f1efa
55
4889e5
4883ec10
48897df8
488975f0
488b4df0
488b45f8
ba00000000
4889ce
4889c7
e8b6fdffff
c9
31d2
31c9
31f6
31ff
c3
\end{lstlisting}  
\end{document}